\documentclass[conference]{IEEEtran}
\usepackage{cite}
\usepackage{amsmath,amssymb,amsfonts}
\usepackage{algorithmic}
\usepackage{graphicx}
\usepackage{textcomp}
\usepackage{xcolor}
\usepackage{booktabs} 
\usepackage{ragged2e}
\usepackage{etoolbox}
\usepackage{multirow}
\usepackage{hyperref}
\usepackage{color}
\usepackage{placeins}
\usepackage{xcolor} 
\usepackage{tikz}

\newcommand\copyrighttext{%
  \footnotesize \textcopyright 2025 IEEE. Personal use of this material is permitted.  Permission from IEEE must be obtained for all other uses, in any current or future media, including reprinting/republishing this material for advertising or promotional purposes, creating new collective works, for resale or redistribution to servers or lists, or reuse of any copyrighted component of this work in other works.}
\newcommand\copyrightnotice{%
\begin{tikzpicture}[remember picture,overlay]
\node[anchor=south,yshift=10pt] at (current page.south) 
  {\fbox{\parbox{\dimexpr\textwidth-\fboxsep-\fboxrule\relax}{\copyrighttext}}};
\end{tikzpicture}%
}

\newcommand\citationnote{%
\footnotesize Accepted version is published in International Joint Conference on Neural Networks, 2025. DOI: \href{https://doi.org/10.1109/IJCNN64981.2025.11229362}{https://doi.org/10.1109/IJCNN64981.2025.11229362}.}
\newcommand\citationnoticebox{%
\begin{tikzpicture}[remember picture,overlay]
\node[anchor=north,yshift=-10pt] at (current page.north)
  {\fbox{\parbox{\dimexpr\textwidth-\fboxsep-\fboxrule\relax}{\citationnote}}};
\end{tikzpicture}%
}

\def\BibTeX{{\rm B\kern-.05em{\sc i\kern-.025em b}\kern-.08em
    T\kern-.1667em\lower.7ex\hbox{E}\kern-.125emX}}
\begin{document}

\title{Effective Graph and Rank-based Contextual Embeddings for Textual and Multimedia Data}

\author{
Thiago César Castilho Almeida\textsuperscript{1}, 
Gustavo Rosseto Letício\textsuperscript{1}, 
Lucas Pascotti Valem\textsuperscript{2}, \\
André Freitas\textsuperscript{3, 4}, 
Daniel Carlos Guimarães Pedronette\textsuperscript{1} \\

\textsuperscript{1} Department of Statistics, Applied Math. and Computing, State University of São Paulo (UNESP), Rio Claro, Brazil \\
\textsuperscript{2} Institute of Mathematical Sciences and Computing, University of São Paulo (USP), São Carlos, Brazil \\
\textsuperscript{3} Department of Computer Science, University of Manchester, Manchester, United Kingdom \\
\textsuperscript{4} Idiap Research Institute, Martigny, Switzerland \\
\{tc.almeida, gustavo.leticio, daniel.pedronette\}@unesp.br, lucas@icmc.usp.br, andre.freitas@manchester.ac.uk
}

\maketitle
\citationnoticebox
\copyrightnotice
\begin{abstract}

In a data-driven world, efficiently organizing and mapping relationships between objects is crucial. Graphs are powerful tools for modeling these connections, being widely used in social networks, telecommunications, and biology. However, graph-based methods often face high computational costs, particularly in memory and space usage. To address this, graph embedding techniques, also referred to as Network Representation Learning, encode graph information into lower-dimensional representations while preserving structural aspects. Traditional methods, however, lack interpretable dimensions.
RaDE (Rank Diffusion Embedding) introduces a new approach using rank-based information, with a key step being the selection of a representative subset of nodes to provide interpretability for its dimensions and improve retrieval tasks. Despite its potential, RaDE's original proposal did not fully explore the effectiveness of representative subset selection across different classes or evaluate embeddings in tasks like classification and clustering.
Inspired by RaDE, this work introduces GRaCE (Graph and Rank-based Contextual Embeddings), a fully unsupervised framework that generates interpretable embeddings by leveraging robust rank-based measures for representative subset selection and node embedding. GRaCE surpasses RaDE and Original Features across diverse datasets, including textual and image collections, excelling in retrieval, classification, and clustering tasks, considering state-of-the-art Transformer models as feature descriptors and Graph Convolutional Networks models in classification tasks.

\end{abstract}


\section{Introduction}

The vast amount of data available in an increasingly globalized and connected world has led to the need for techniques capable of organizing it efficiently, allowing the storage of relationships between elements. In this context, the representation of information using Graphs has proven to be an excellent alternative, effectively representing relationships between individuals in an easily interpretable way. This approach has been widely used in various scenarios, such as social networks, molecular structures, biological protein interaction networks, and recommendation systems \cite{survey2018Hamilton}.

Many machine learning techniques leverage graph structures for tasks like classification, link prediction, clustering, and visualization \cite{survey2018Goya}. However, graph-based operations often face high computational costs in terms of time and space \cite{survey2018Cai}. To address these challenges, graph embedding techniques, also known as Network Representation Learning (NRL) \cite{Cui2019NRL}, have emerged as effective solutions.

Graph embedding approaches aim to encode the structural information of the graph into a low-dimensional vector representation, enabling the efficient application of classification, clustering, or visualization methods. The main goal of these techniques is to optimize the computational costs associated with graph-based methods while preserving some structural information of the network \cite{survey2018Cai}. As a result, vector representations can maintain or even improve the results obtained in machine learning tasks \cite{rade2022Filipe}.

Graph embedding methods can be organized along several axes \cite{survey2018Goya}. However, like many classical machine learning techniques, including deep learning models for extracting feature representations from text and multimedia, approaches such as DeepWalk \cite{deepwalk} and node2vec \cite{node2vec}, as well as other deep learning–based graph embedding methods, are often regarded as ``black boxes''. This implies that, despite their ability to effectively encode graph structure, the underlying reasons for their success remain insufficiently understood \cite{interpretability1028Liu}: each embedding dimension lacks explicit semantic meaning, and the optimization procedure is largely opaque. This interpretability gap becomes a significant barrier in domains where understanding and trusting the learned representations is as important as their predictive power, such as in medical and legal applications. Without interpretability, reliance on these black box methods can compromise user trust \cite{interpretability1028Liu}. To address this challenge, recent research has focused on both measuring interpretability \cite{interpretability1028Liu} and designing graph embedding techniques with semantical dimensions \cite{rade2022Filipe,dine}.

Dimension‐based Interpretable Node Embedding (DINE) \cite{dine} tackles this by applying a post‐processing step to any existing node embedding: it projects nodes into a lower‐dimensional space where each dimension corresponds to human‐understandable structural patterns and substructures within the input graph.

Likewise, RaDE (Rank Diffusion Embedding) \cite{rade2022Filipe} aims to produce semantic dimensions by building embeddings through three sequential stages: the construction of a rank-based similarity graph; the selection of highly effective representative nodes; and node embedding, where each node is embedded into a new vector space defined by its diffusion‐based affinities to these representatives, ensuring that each embedding dimension have individual meaning.

In RaDE, selecting a small representative subset of nodes is a critical step. This challenge is formally referred to as the Representative Selection task \cite{rsgnn2023Kazemi}. A good representative subset should encapsulate as much information from the original set as possible while maintaining low redundancy among the selected elements \cite{measures2005Feng}. The Representative Selection task has many applications, like summarization, active learning, data compression, model training cost reduction, and many other domains \cite{rsgnn2023Kazemi}. Recent works are proposing methods that are capable of extracting a representative subset from huge datasets \cite{rsgnn2023Kazemi,measures2005Feng}, and the representative subsets are being applied in scenarios, such as active learning \cite{activeFramework2017Cai}, identifying the best elements to label, and semi-supervised learning \cite{grain2021Zhang}, selecting only a small subset to be annotated.

This work introduces a novel framework named \textbf{GRaCE} (Graph and Rank-based Contextual Embeddings), which aims to generate more informative node embeddings by leveraging representative subsets selected through rank-based measures. GRaCE is inspired by the principles of RaDE but extends them by integrating more robust and recent rank-based measures, including rank correlation measures such as JaccardMax \cite{jacmax} and Reciprocal \cite{reciprocalcorrelation2014Pedronette}, which estimate the similarity between two nodes based on their neighborhood. Additionally, GRaCE employs unsupervised effectiveness estimation measures, also known as Query Performance Prediction measures, which assess the quality of a query or ranking without requiring labeled data. Examples include Reciprocal Density \cite{reciprocalestimation2014Pedronette} and a newly proposed Accumulated JaccardMax \cite{accjacmax2024Almeida}, which uses rank correlation to estimate query effectiveness.

While the effectiveness of representative selection was not evaluated in the original RaDE work \cite{rade2022Filipe}, GRaCE provides a broader contribution: it not only offers an evaluation of this component but also proposes a complete framework that enhances representation learning through advanced rank-based strategies.
The hypothesis is that incorporating these advanced and recent rank-based measures will enhance the effectiveness of the representative subset selection.
The selected subsets are expected to yield more effective embeddings, improving performance in tasks like classification, clustering, and retrieval.

Thus, the main contributions of this paper can be summarized as follows: \emph{\textbf{(i)}} The introduction of a new framework named GRaCE which uses robust rank-based measures for representative subset selection and low-dimensional vector representation creation; \emph{\textbf{(ii)}} An evaluation of the representativeness of the subset selection in RaDE and GRaCE; \emph{\textbf{(iii)}} An evaluation of the embeddings produced by RaDE and GRaCE in retrieval, as well as in other machine learning tasks like classification and clustering, which were not examined in \cite{rade2022Filipe}.

This paper is organized as follows. Section \ref{sec:background} formally defines the embedding and rank problem. Section \ref{sec:rade} describes GRaCE framework, explaining its key steps and the rank-based measures. Section \ref{sec:experimental-evaluation} presents the experimental evaluation, detailing the datasets, the experimental protocol, and the results obtained. Section \ref{sec:conclusion} concludes the work.

\section{Formal Definition}
\label{sec:background}
This section formally presents the formal definition of the problem in this work, including the graph embedding model and the rank model, following the notation used in \cite{rade2022Filipe}.

\subsubsection{Graph Embedding} Let $\mathcal{C} = \{e_1, e_2, \dots, e_n\}$ be a collection, $n = |\mathcal{C}|$, which can be represented by a graph $\mathcal{G}(\mathcal{V}, \mathcal{E})$, where $\mathcal{V}$ is the node set, and $\mathcal{E}$ denotes the edge set. Every $e_j \in C$ is a node in the set $\mathcal{V}$. If $(e_j, e_i) \in \mathcal{E}$, we can say that the vertices representing the elements $e_j$ and $e_i$ are connected in the graph. Weights can be assigned to the edges, represented by an adjacency matrix $\mathcal{S}$, such that the weight of an edge $(e_i, e_j)$ is given by $s_{i,j}$. 

A graph embedding task, specifically node embedding, can be defined as a function \( f : \mathcal{V} \rightarrow \mathbb{R}^d \), where \( d \ll |\mathcal{V}| \). This function preserves the graph's structural information, ensuring that nodes close in \(\mathcal{G}\) remain close in the \(\mathbb{R}^d\) space.

For multimedia data, generating a similarity graph is crucial for graph embedding. This involves extracting feature vectors, computing Euclidean-like distances between elements, and using these distances to determine similarity values.

\subsubsection{Rank Model} Ranking tasks are typically defined using pairwise dissimilarity measures, where the distance between two elements $e_i$ and $e_j$ is represented by $\rho(i, j)$. Let $e_q$ be a query element. A ranked list $\tau_q = (e_1, e_2, \dots, e_n) $ is a permutation of the collection $\mathcal{C}$. The rank of an element $e_i$ in the ranked list $\tau_q$ is defined as $\tau_q(i)$. If $\tau_q(i) < \tau_q(j)$, then $\rho(q, i) \leq \rho(q, j)$, meaning that the distance between $e_q$ and $e_i$ is less than the distance between $e_q$ and $e_j$, guaranteeing the most similar objects to the query be on the top positions. Every element in the collection has its own ranked list. The set of all ranked lists can be defined as $\mathcal{T}$. The neighborhood set that contains the most $k$ similar elements to a query $e_q$ can be defined as $\mathcal{N}(q, k)$.

\section{GRaCE: Contextual Embeddings}
\label{sec:rade}

\begin{figure*}[!ht]
    \centering
    \includegraphics[width=0.9\linewidth]{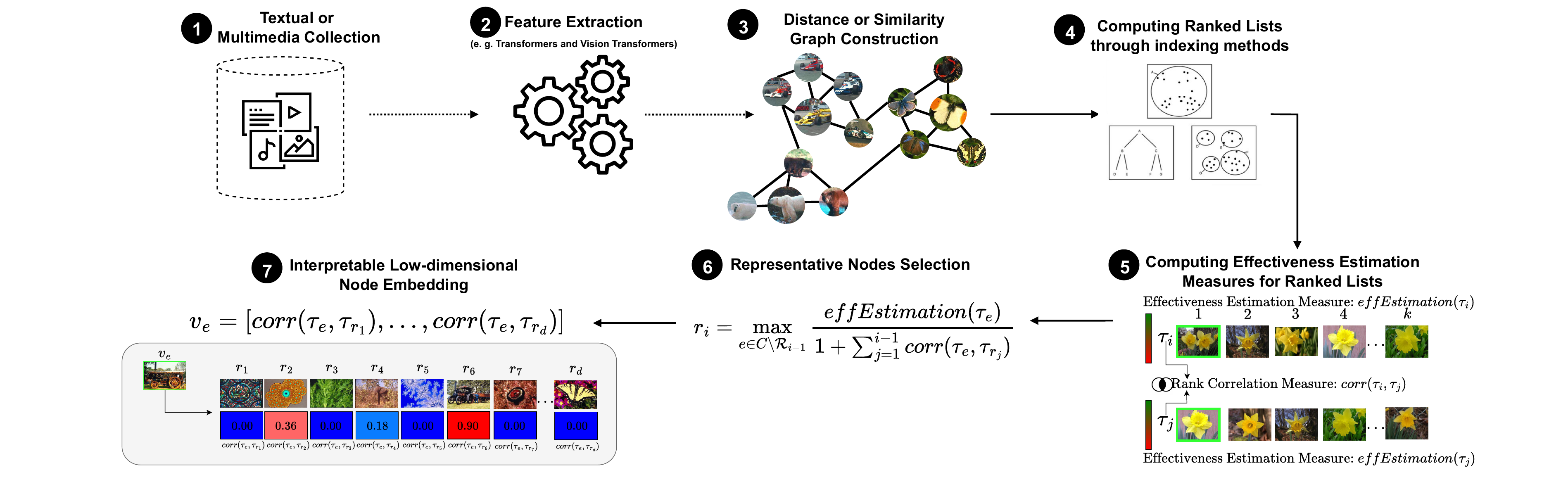}
    \vspace{-5mm}
    \caption{Main steps of GRaCE applied to textual and multimedia data.}
    \label{fig:proposed-approach}
\end{figure*}

This section introduces GRaCE, a novel graph embedding technique designed to generate low-dimensional, semantically meaningful representations of data. Figure \ref{fig:proposed-approach} illustrates the use of GRaCE in textual and multimedia data. GRaCE leverages rank-based measures to identify representative nodes and construct new vector representations that preserve the structural properties of the original data. It follows three main steps:

    \emph{\textbf{A.}} \textbf{Rank-based Similarity Graph Construction:} The input graph is transformed into a $k$-nearest neighbors (kNN) graph, where each node is connected to its $k$-nearest neighbors based on similarity metrics. This step ensures that the graph captures local relationships and structural patterns within the data;
    
    \emph{\textbf{B.}} \textbf{Representative Subset Selection:} A subset of nodes is selected to serve as representatives of the entire graph. These nodes are chosen to encode as much information as possible from the original data, prioritizing highly effective nodes while minimizing redundancy. This step is crucial since identifying good or bad representative nodes can directly affect next step;
    
    \emph{\textbf{C.}} \textbf{Node Embedding:} Each node in the graph is assigned to a new vector representation based on its similarity to the representative nodes selected in the previous step.

By combining these steps, GRaCE creates semantically rich embeddings, and suitable for a wide range of applications, including clustering, classification, and visualization, reducing computational costs or even improving the effectiveness of these tasks. The following subsections detail its main steps.

\subsection{Rank-based Similarity Graph Construction}

The proposed framework begins by building a kNN graph, where each node is connected to its $k$-nearest neighbors using weighted edges or feature-based distances (e.g., Euclidean) suited to textual or multimedia data. This structure enables the use of rank-based measures focused on local relationships, effectively capturing the data's underlying structure.

\subsection{Representative Subset Selection}

Identifying representative nodes is crucial for the effectiveness of the produced embeddings, capturing as much information as possible from the original dataset's structure while minimizing redundancy within the selected subset. To ensure the selection of highly effective representative nodes, elements with the highest effectiveness estimation are prioritized. However, elements that are highly correlated with previously selected nodes are penalized to avoid redundancy.
Thus, to construct a highly effective representative set $\mathcal{R}$, each element $r_i \in \mathcal{R}$ is selected based on the following criteria: \emph{1. Effectiveness Estimation (QPP):} Prioritize nodes with the highest effectiveness scores, ensuring they capture significant structural information; \emph{2. Correlation Penalization:} Penalize nodes that are highly correlated with already selected representatives to maintain diversity and reduce redundancy.
Formally, each element $r_i$ can be selected using the equation: 
\begin{equation}
    r_i = \max_{e \in C \setminus \mathcal{R}_{i-1}} \frac{effEstimation(\tau_e)}{1 + \sum_{j=1}^{i-1} corr(\tau_e,\tau_{r_j})}.
\end{equation}

This approach ensures that the representative set $\mathcal{R}$ is highly effective and non-redundant, leading to more effective embeddings that preserve the structure of the original space. Rank correlation and unsupervised effectiveness estimation measures will be presented as follows.

\subsubsection{\textbf{Rank Correlation Measures}}
\label{subsec:rank-correlation}

Rank correlation measures are used to determine the similarity between two ranked lists. Formally, a rank correlation metric can be defined as a function $corr: \mathcal{T} \times \mathcal{T} \to \mathbb{R}$, assigning a value between 0 (completely different rankings) and 1 (identical rankings). Rank correlation measures usually evaluate the similarity between the ranked lists, considering only the top-$k$ positions, producing a $\mathcal{N}(q, k)$ set for the ranked list of an element $e_q$. This work employed two similarity measures, Reciprocal and JaccardMax, which have shown greater effectiveness in rank-based tasks compared to traditional measures like Jaccard and RBO \cite{rbo}, as demonstrated in prior studies \cite{reciprocalcorrelation2014Pedronette,jacmax}.

The \textbf{Reciprocal} measure \cite{reciprocalcorrelation2014Pedronette} computes the number of reciprocal neighbors at the top positions of ranked lists, assigning a weight proportionally to their ranks.
    \begin{equation}\label{eq:recdistance}
    \scalebox{0.97}{
    $ n_r(\tau_q, \tau_i, k) = \frac{\sum_{j \in \mathcal{N}(q, k)} \sum_{l \in \mathcal{N}(i, k)} f_r(j, l) \times w_r(q, j) \times w_r(i, l)}{k^4},$
 }
 \end{equation}

 \noindent where $f_r$ determines if a pair of elements $(e_j, e_l)$ are reciprocal neighbors, and $w_r(q,j) = k + 1 - \tau_q(j)$ is a weight function.

The \textbf{JaccardMax (JacMax)} \cite{jacmax} measure considers the highest Jaccard index value obtained until a depth $k$ as the similarity value between two ranked lists.
 \begin{equation}\label{eq:jaccardmax}
JacMax(\tau_i,\tau_j,k) = \max\limits_{1 \leq k_d \leq k} \dfrac{| \mathcal{N}(i,k_d)  \cap \mathcal{N}(j,k_d) |}{| \mathcal{N}(i,k_d) \cup \mathcal{N}(j,k_d) |}.
\end{equation}

\subsubsection{\textbf{Effectiveness Estimation Measures}}
\label{subsec:effectivenes-estimation}

Also referred to in the literature as Query Performance Prediction (QPP), effectiveness estimation measures can be defined as a function $effEstimation: \mathcal{T} \to \mathbb{R}$ and are responsible for evaluating a ranked list's effectiveness, distinguishing good and bad queries in an unsupervised fashion by considering the contextual information between the ranked lists to compute their effectiveness. The measures used in this work are defined below.

The \textbf{Reciprocal Neighborhood Density (Rec. Density)} score \cite{reciprocalestimation2014Pedronette} is based on the Reciprocal \cite{reciprocalcorrelation2014Pedronette} measure, but as we are interested in computing the number of reciprocal neighbors of a query ranked list, we define the score considering the same ranked list in both inputs in Equation \ref{eq:recdistance}. Formally, $R_d(q) = n_r(\tau_q,\tau_q, k)$.

The \textbf{Accumulated JaccardMax (Acc. JacMax)} \cite{accjacmax2024Almeida} can be understood as the average of similarity, given by the correlation measure JaccardMax \cite{jacmax}, between the query-ranked list and the rankings of its $k$-nearest neighbors. 
\begin{equation}
\scalebox{0.95}{
$AccJacMax(q, k, \alpha) = \frac{\sum_{j \in \mathcal{N}(q, k)}JacMax(\tau_q, \tau_j, k) \times \alpha^{\tau_q(j)}}{k},$
}\end{equation}
 where $0 < \alpha \leq 1$. The higher the $\alpha$, the greater the weight assigned to the top positions.

\subsection{Node Embedding}

Since the representative subset $\mathcal{R}$ is selected, for every element in the dataset, a new low-dimensional semantically meaningful representation is created considering the similarity of the element with each of the representative nodes. In RaDE \cite{rade2022Filipe}, the idea of the similarity between two nodes $e_i$ and $e_j$ is defined by the affinity between them, given by the $A$ diffusion matrix in the position $a_{i, j}$. In GRaCE, the idea of similarity is given by a rank correlation measure between the ranked lists of $e_i$ and $e_j$. Formally, for each element $e$ from the collection $\mathcal{C}$, the new embedding is created by: $v_e = [corr(\tau_e, \tau_{r_1}), \dots, corr(\tau_e, \tau_{r_d})].$

\section{Experimental Evaluation}
\label{sec:experimental-evaluation} 
This section presents the experimental evaluation conducted in this work.
Subsection \ref{subsec:datasets} presents the datasets and feature descriptors. Subsection \ref{subsec:exp-protocol} outlines the experimental protocol, including evaluation measures, algorithms, and parameter settings.
Subsections \ref{subsec:subset-selection}, \ref{subsec:retrieval-tasks}, \ref{subsec:classification-tasks}, and \ref{subsec:clustering-tasks}, evaluate GRaCE performance across four tasks: representative subset selection, retrieval, classification, and clustering, respectively. 

\subsection{Datasets and Representations}
\label{subsec:datasets}

The evaluation was conducted on four datasets of two modalities (image and text): Flowers~\cite{flowers} (1,360 images, 17 classes), Corel5k~\cite{corel5k} (5,000 images, 50 classes), BBCNews~\cite{bbcnews} (2,225 articles, 5 categories), and WOS-5736~\cite{kowsari2017WOS5736} (5,736 abstracts, 11 disciplines).
For the image datasets, three Vision Transformer (ViT) models were used: ViT-B16 \cite{VIT} and Swin-Tf \cite{swin}, both pre-trained on ImageNet \cite{imagenet}, and DINOv2-B14, a self-supervised ViT model trained on the large-scale unlabeled LVD-142M dataset \cite{oquab2024dinov2}. In textual data, features were extracted by pre-trained Sentence Transformers \cite{reimers-2019-sentence-bert} models available on Hugging Face \href{https://huggingface.co/models}{(https://huggingface.co/models)}: \textit{nthakur/mcontriever-base-msmarco} \cite{mcontriever2022Gautier}, \textit{sentence-transformers/all-MiniLM-L6-v2} \cite{wang2020minilmdeepselfattentiondistillation}, and \textit{sentence-transformers/all-distilroberta-v1} \cite{Sanh2019DistilBERTAD}.

\subsection{Experimental Protocol}
\label{subsec:exp-protocol}

This work evaluates the proposed method across multiple tasks, including classification, representative selection, retrieval, and clustering. While classification relied on labeled data, the other tasks were unsupervised and evaluated using appropriate ground-truth-based performance measures.

To evaluate the task of highly effective representative subset selection, two measures based on \cite{measures2005Feng} were adapted to our use. The first is called Coverage and measures the percentage of the dataset classes covered by the selected representative subset.
The second measure is named Clustering Accuracy (Accuracy), which counts the number of elements that have the same class label as the most similar representative node. 

Retrieval tasks were evaluated using Precision at depths 2, 4, and 16, along with Mean Average Precision (MAP) at depth 1000. Classification performance was measured using Accuracy, suitable for balanced classes. Clustering was assessed with Normalized Mutual Information (NMI), which quantifies shared information between true labels and predicted clusters.

For all tasks evaluated in this study, the generated embeddings were composed of 128 dimensions ($d = 128$). This dimensionality was chosen to align with the value used in \cite{rade2022Filipe}. Additionally, the $t$ parameter of RaDE was set to $t = 2$, following the same configuration as its publication. This ensures consistency and facilitates a fair evaluation of the proposed approach against the baseline.

The parameter \( L \) in RaDE corresponds to the \( k \) parameter in GRaCE for constructing the similarity graph and computing rank-based measures. For the Flowers and Corel5k datasets, values were set to $70$ and $100$, respectively, following \cite{rade2022Filipe}. For BBCNews and WOS-5736, which were not in the original study, \( k \) was set to $200$ based on the dataset structure. The Accumulated JaccardMax \( \alpha \) parameter was set to $0.95$, based on the best overall results from \cite{accjacmax2024Almeida}.

In classification tasks, the methods were evaluated following the protocol used in \cite{valem2023cviu}, considering the average Accuracy between five rounds of 10-fold cross-validation, using 1 fold for training, and the others for testing. Four methods were evaluated. The KNN (K-Nearest Neighborhoods) \cite{cover1967knn} is a classification method that determines the label of an element based on the majority label of its $k$-nearest neighborhoods, with $k = 10$ in all executions. Three Graph Convolutional Network (GCN) models were considered, which are the GCN-NET \cite{kipf2017gcn}, the first model introduced in the literature, the Simple Graph Convolution (SGC) \cite{wu2019sgc}, a simplified version of the GCN-NET model, and the Approximate Personalized Propagation of Neural Predictions (APPNP) model \cite{PaperAPPNP_2019}, which combines a GCN with the PageRank algorithm. For all GCN models and datasets, the ranked lists were used to construct a kNN graph, with $k=40$, using the Adam optimizer with a learning rate of $10^{-4}$ and trained for $200$ epochs.

In clustering tasks, four algorithms were evaluated. KMeans \cite{lloyd1982kmeans} assigns points to the nearest cluster, adjusting centroids iteratively, with the number of clusters set to the dataset's class count. Affinity Propagation (AP) \cite{affinityProp2007} identifies clusters via message passing without requiring a predefined cluster count. Agglomerative Hierarchical Clustering (AHC) \cite{murtagh1983agglomerative} builds a cluster hierarchy by merging the closest pairs, with the number of clusters determined by cutting the dendrogram at a specific level, matching the dataset's class count. HDBSCAN \cite{hdbscan} extends DBSCAN by using hierarchical and density-based methods to group points based on local density.

Except for the GCN models, all the experiments for classification and clustering tasks were conducted in the Python programming language, using the \textit{scikit-learn} library (\href{https://scikit-learn.org/stable/}{https://scikit-learn.org/stable/}). The omitted parameters were set to the library default.

\subsection{Representative Nodes Selection Task}
\label{subsec:subset-selection}

\setlength{\tabcolsep}{2pt}
\begin{table}[t!]
    \centering
        \caption{Evaluation of representative subset effectiveness.}
        \label{tab:selection}
        \vspace{-2mm}
\resizebox{\columnwidth}{!}{
    \begin{tabular}{c|c|c|c|c|cc}
    \hline
         \textbf{Dataset} & \textbf{Descriptor} & \multicolumn{2}{c|}{\textbf{QPP}} & \textbf{Correlation} & \textbf{Coverage} & \textbf{Accuracy}\\
         \hline
          \multirow{15}{*}{\rotatebox[origin=c]{45}{\textbf{Flowers}}}&\multirow{5}{*}{\rotatebox[origin=c]{45}{\textbf{ViT-B16}}}&  \multicolumn{3}{c|}{RaDE} & \textbf{1}&\textbf{0.9779} \\
         \cline{3-5} 
           && \multirow{4}{*}{\rotatebox[origin=c]{90}{\textbf{GRaCE}}} & Rec. Density & Reciprocal & \textbf{1}&0.9721\\
           &&& Rec. Density & JacMax & \textbf{1}&0.9735\\
           & &&Acc. JacMax & Reciprocal & \textbf{1}&0.9713\\
          && &Acc. JacMax & JacMax & \textbf{1}&0.9706\\
          \cline{2-7}
          &\multirow{5}{*}{\rotatebox[origin=c]{45}{\textbf{Swin-Tf}}}&  \multicolumn{3}{c|}{RaDE} & \textbf{1}&0.9743 \\
         \cline{3-5}
           
           && \multirow{4}{*}{\rotatebox[origin=c]{90}{\textbf{GRaCE}}} &Rec. Density & Reciprocal & 0.9412&0.9397\\
           &&& Rec. Density & JacMax & \textbf{1}&\textbf{0.9787}\\
           & &&Acc. JacMax & Reciprocal & \textbf{1}&0.9515\\
          & &&Acc. JacMax & JacMax & \textbf{1}&0.9824\\
          \cline{2-7}
          &\multirow{5}{*}{\rotatebox[origin=c]{45}{\textbf{DINOv2-B14}}}& \multicolumn{3}{c|}{RaDE} & \textbf{1}&0.9978 \\
         \cline{3-5}
           
           && \multirow{4}{*}{\rotatebox[origin=c]{90}{\textbf{GRaCE}}} &Rec. Density & Reciprocal & \textbf{1}&0.9963\\
           &&& Rec. Density & JacMax & \textbf{1}&0.9963\\
           &&& Acc. JacMax & Reciprocal & \textbf{1}&0.9963\\
          & &&Acc. JacMax & JacMax & \textbf{1}&\textbf{1}\\
         \hline
          \multirow{15}{*}{\rotatebox[origin=c]{45}{\textbf{Corel5k}}} & \multirow{5}{*}{\rotatebox[origin=c]{45}{\textbf{ViT-B16}}}&  \multicolumn{3}{c|}{RaDE} & \textbf{1}&\textbf{0.8988} \\
         \cline{3-5}
           
           && \multirow{4}{*}{\rotatebox[origin=c]{90}{\textbf{GRaCE}}} &Rec. Density & Reciprocal & 0.94&0.8494\\
           && &Rec. Density & JacMax & 0.98&0.878\\
           && &Acc. JacMax & Reciprocal & 0.90&0.8338\\
          && &Acc. JacMax & JacMax & 0.98&0.8900\\
          \cline{2-7}
          & \multirow{5}{*}{\rotatebox[origin=c]{45}{\textbf{Swin-Tf}}}&  \multicolumn{3}{c|}{RaDE} & \textbf{0.94}&0.8912 \\
         \cline{3-5}
           
           && \multirow{4}{*}{\rotatebox[origin=c]{90}{\textbf{GRaCE}}} &Rec. Density & Reciprocal & 0.84&0.7954\\
           &&& Rec. Density & JacMax & 0.92&0.8796\\
           && &Acc. JacMax & Reciprocal & 0.82&0.7862\\
          && &Acc. JacMax & JacMax & \textbf{0.94}&\textbf{0.8916}\\
          \cline{2-7}
          & \multirow{5}{*}{\rotatebox[origin=c]{45}{\textbf{DINOv2-B14}}}& \multicolumn{3}{c|}{RaDE} & 0.96&0.9064 \\
         \cline{3-5}
           
           && \multirow{4}{*}{\rotatebox[origin=c]{90}{\textbf{GRaCE}}} &Rec. Density & Reciprocal & 0.92&0.8772\\
           && &Rec. Density & JacMax & 0.96&0.9122\\
           && &Acc. JacMax & Reciprocal & 0.94&0.8928\\
          && &Acc. JacMax & JacMax & \textbf{0.98}&\textbf{0.9340}\\
          \hline
          \multirow{15}{*}{\rotatebox[origin=c]{45}{\textbf{BBCNews}}} & \multirow{5}{*}{\rotatebox[origin=c]{45}{\textbf{mContriever}}}& \multicolumn{3}{c|}{RaDE} & \textbf{1} & 0.8418\\
         \cline{3-5}
          & & \multirow{4}{*}{\rotatebox[origin=c]{90}{\textbf{GRaCE}}} &Rec. Density & Reciprocal & \textbf{1} & 0.8022\\
           && &Rec. Density & JacMax & \textbf{1}&0.8387\\
           && &Acc. JacMax & Reciprocal & \textbf{1} & 0.7955\\
          && &Acc. JacMax & JacMax & \textbf{1}&\textbf{0.8598}\\
          \cline{2-7}
          &\multirow{5}{*}{\rotatebox[origin=c]{45}{\textbf{DistilRoBERTa}}}& \multicolumn{3}{c|}{RaDE} & \textbf{1} & 0.8440\\
         \cline{3-5}
           && \multirow{4}{*}{\rotatebox[origin=c]{90}{\textbf{GRaCE}}} &Rec. Density & Reciprocal & \textbf{1}&0.8427\\
           && &Rec. Density & JacMax &\textbf{1}&\textbf{0.8535}\\
           && &Acc. JacMax & Reciprocal & \textbf{1}&0.7474\\
          && &Acc. JacMax & JacMax &\textbf{1}& 0.8440\\
          \cline{2-7}
          &\multirow{5}{*}{\rotatebox[origin=c]{45}{\textbf{MiniLM}}}&  \multicolumn{3}{c|}{RaDE} & \textbf{1}& \textbf{0.8490} \\
         \cline{3-5}
           && \multirow{4}{*}{\rotatebox[origin=c]{90}{\textbf{GRaCE}}} &Rec. Density & Reciprocal & \textbf{1}&0.8283\\
           && &Rec. Density & JacMax & \textbf{1}&0.8463\\
           && &Acc. JacMax & Reciprocal & \textbf{1}&0.8108\\
          && &Acc. JacMax & JacMax & \textbf{1}&0.8373\\
          \hline
          \multirow{15}{*}{\rotatebox[origin=c]{45}{\textbf{WOS-5736}}} & \multirow{5}{*}{\rotatebox[origin=c]{45}{\textbf{mContriever}}}&  \multicolumn{3}{c|}{RaDE} &0.9091  & 0.6168\\
         \cline{3-5}
           && \multirow{4}{*}{\rotatebox[origin=c]{90}{\textbf{GRaCE}}} &Rec. Density & Reciprocal & 0.9091&0.6159\\
           &&& Rec. Density & JacMax & \textbf{1}&0.6592\\
           && &Acc. JacMax & Reciprocal & 0.9091&0.6212\\
          && &Acc. JacMax & JacMax &\textbf{1} &\textbf{0.6604}\\
          \cline{2-7}
          &\multirow{5}{*}{\rotatebox[origin=c]{45}{\textbf{DistilRoBERTa}}}& \multicolumn{3}{c|}{RaDE} & \textbf{1} & 0.6119\\
         \cline{3-5}
           && \multirow{4}{*}{\rotatebox[origin=c]{90}{\textbf{GRaCE}}} &Rec. Density & Reciprocal & \textbf{1}&0.6128\\
           &&& Rec. Density & JacMax & \textbf{1}&\textbf{0.6478}\\
           && &Acc. JacMax & Reciprocal & \textbf{1}&0.5962\\
          & &&Acc. JacMax & JacMax & \textbf{1}&0.6412\\
          \cline{2-7}
          &\multirow{5}{*}{\rotatebox[origin=c]{45}{\textbf{MiniLM}}}& \multicolumn{3}{c|}{RaDE} & \textbf{1} & 0.6288\\
         \cline{3-5}
           && \multirow{4}{*}{\rotatebox[origin=c]{90}{\textbf{GRaCE}}} &Rec. Density & Reciprocal & \textbf{1}&0.6161\\
           && &Rec. Density & JacMax & \textbf{1}&0.6482\\
           &&& Acc. JacMax & Reciprocal & \textbf{1}&0.6133\\
          & &&Acc. JacMax & JacMax & \textbf{1}&\textbf{0.6538}\\
          \hline
    \end{tabular}
    }
\end{table}
Table \ref{tab:selection} presents the results of the representative subset effectiveness, where GRaCE consistently outperforms the RaDE baseline, achieving the best values of evaluation measures in at least two out of three feature descriptors across all datasets.

As shown in Table \ref{tab:selection}, the Coverage values for both RaDE and GRaCE are consistently high across all cases, with only slight differences between them, and in every scenario, both methods successfully encompass the majority of classes. However, GRaCE’s variants that employ JaccardMax as the rank correlation measure tend to yield superior evaluation results. For example, when using the DINOv2-B14 model, GRaCE with JaccardMax consistently achieves near-perfect Clustering Accuracy, reaching 1.0 on the Flowers dataset and 0.9340 on Corel5k; and, on the textual WOS-5736 dataset, it surpasses RaDE’s Clustering Accuracy by roughly 3\% across all feature descriptors. These findings underscore GRaCE’s enhanced ability to select more representative subsets than RaDE, demonstrating its robustness and reliability across both structured image data and diverse text corpora.

\subsection{Retrieval Tasks}
\label{subsec:retrieval-tasks}

\begin{table}[t!]
    \centering
        \caption{Evaluation of retrieval tasks across all datasets.}
\vspace{-2mm}
    \resizebox{.50\textwidth}{!}{
    \begin{tabular}{c|c|c|c|c|cccc}
    \hline
         \textbf{Dataset} & \textbf{Descriptor} &\multicolumn{2}{c|}{\textbf{QPP}} & \textbf{Correlation} & \textbf{P@2} & \textbf{P@4} & \textbf{P@16} & \textbf{MAP}\\
          \hline
          \multirow{18}{*}{\rotatebox[origin=c]{45}{\textbf{Flowers}}} & \multirow{5}{*}{\rotatebox[origin=c]{45}{\textbf{ViT-B16}}}& \multicolumn{3}{c|}{Original Features} &\textbf{0.9934}&	\textbf{0.9879}	&\textbf{0.9731}	&0.8752\\
         \cline{3-5}
         && \multicolumn{3}{c|}{RaDE} &0.9904&	0.9836&\textbf{0.9731}	&0.9675\\
         \cline{3-5}
          & & \multirow{4}{*}{\rotatebox[origin=c]{90}{\textbf{GRaCE}}}& Rec. Density & Reciprocal & 0.9890&0.9814&	0.9719	&\textbf{0.9692}\\
           &&& Rec. Density & JacMax & 0.9871&	0.9783	&0.9648	&0.9565\\
           &&& Acc. JacMax & Reciprocal &0.9890&	0.9812	&0.9715	&0.9691\\
          & &&Acc. JacMax & JacMax &0.9890&0.9796	&0.9663	&0.9567\\
          \cline{2-9}
         &\multirow{5}{*}{\rotatebox[origin=c]{45}{\textbf{Swin-Tf}}}& \multicolumn{3}{c|}{Original Features} &\textbf{0.9982}&	0.9939&	0.9816	&0.9291\\
         \cline{3-5}
         && \multicolumn{3}{c|}{RaDE} &0.9978&	0.9961	&0.9932&	0.9902\\
         \cline{3-5}
           && \multirow{4}{*}{\rotatebox[origin=c]{90}{\textbf{GRaCE}}}& Rec. Density & Reciprocal &\textbf{0.9982}&	0.9960	&0.9939&	0.9878\\
           &&& Rec. Density & JacMax &\textbf{0.9982}&	\textbf{0.9963}	&\textbf{0.9941}	&\textbf{0.9925}\\
           &&& Acc. JacMax & Reciprocal &\textbf{0.9982}&	0.9961	&0.9939	&0.9876\\
          && &Acc. JacMax & JacMax & \textbf{0.9982}	&0.9961	&0.9937	&0.9919\\
          \cline{2-9}
          &\multirow{5}{*}{\rotatebox[origin=c]{45}{\textbf{DINOv2-B14}}}& \multicolumn{3}{c|}{Original Features} &\textbf{1}&	\textbf{1}	&\textbf{0.9989}	&0.9877\\
         \cline{3-5}
         & &\multicolumn{3}{c|}{RaDE} &\textbf{1}&	0.9996&0.9979	&0.9973\\
         \cline{3-5}
           && \multirow{4}{*}{\rotatebox[origin=c]{90}{\textbf{GRaCE}}}& Rec. Density & Reciprocal & 0.9890	&0.9814&	0.9719	&0.9692\\
           & &&Rec. Density & JacMax & \textbf{1}&	0.9996	&0.9975	&0.9967\\
           & &&Acc. JacMax & Reciprocal &\textbf{1}&	0.9996	&0.9977	&0.9976\\
          & &&Acc. JacMax & JacMax &\textbf{1}	&\textbf{1}	&0.9985	&\textbf{0.9981}\\
          \hline
          \multirow{18}{*}{\rotatebox[origin=c]{45}{\textbf{Corel5k}}} & \multirow{5}{*}{\rotatebox[origin=c]{45}{\textbf{ViT-B16}}}& \multicolumn{3}{c|}{Original Features} &\textbf{0.9828}&	\textbf{0.9670}	&0.9214	&0.7502\\
         \cline{3-5}
         && \multicolumn{3}{c|}{RaDE} &0.9695&	0.9502&\textbf{0.9242}	&\textbf{0.8658}\\
         \cline{3-5}
          & & \multirow{4}{*}{\rotatebox[origin=c]{90}{\textbf{GRaCE}}}& Rec. Density & Reciprocal & 0.9621	&0.9368&	0.9070	&0.8533\\
           &&& Rec. Density & JacMax & 0.9592&	0.9381	&0.9037	&0.8397\\
           &&& Acc. JacMax & Reciprocal &0.9611&	0.9372	&0.9044	&0.8478\\
          & &&Acc. JacMax & JacMax &0.9631	&0.9378	&0.9036	&0.8315\\
          \cline{2-9}
         & \multirow{5}{*}{\rotatebox[origin=c]{45}{\textbf{Swin-Tf}}}& \multicolumn{3}{c|}{Original Features} &\textbf{0.9920}&	\textbf{0.9803}&	0.9456	&0.7327\\
         \cline{3-5}
         & &\multicolumn{3}{c|}{RaDE} &0.9863&	0.9750	&\textbf{0.9574}&	\textbf{0.9072}\\
         \cline{3-5}
           && \multirow{4}{*}{\rotatebox[origin=c]{90}{\textbf{GRaCE}}}& Rec. Density & Reciprocal &0.9764&	0.9596	&0.9320&	0.8716\\
           & &&Rec. Density & JacMax &0.9776&	0.9617	&0.9327	&0.8686\\
           &&& Acc. JacMax & Reciprocal &0.9767&	0.9600	&0.9323	&0.8715\\
          & &&Acc. JacMax & JacMax & 0.9801	&0.9664	&0.9412	&0.8718\\
          \cline{2-9}
          &\multirow{6}{*}{\rotatebox[origin=c]{45}{\textbf{DINOv2-B14}}}& \multicolumn{3}{c|}{Original Features} &\textbf{0.9875}&	\textbf{0.9766}	&0.9444	&0.8128\\
         \cline{3-5}
         & &\multicolumn{3}{c|}{RaDE} &0.9829&	0.9698&\textbf{0.9498}	&0.8790\\
         \cline{3-5}
           && \multirow{4}{*}{\rotatebox[origin=c]{90}{\textbf{GRaCE}}}& Rec. Density & Reciprocal & 0.9775	&0.9640&	0.9436	&0.9037\\
           & &&Rec. Density & JacMax & 0.9769&	0.9630	&0.9416	&0.8927\\
           && &Acc. JacMax & Reciprocal &0.9789&	0.9654	&0.9466	&\textbf{0.9070}\\
          & &&Acc. JacMax & JacMax &0.9792	&0.9662	&0.9445	&0.8868\\
          \hline
         \multirow{18}{*}{\rotatebox[origin=c]{45}{\textbf{BBCNews}}} & \multirow{5}{*}{\rotatebox[origin=c]{45}{\textbf{mContriever}}}& \multicolumn{3}{c|}{Original Features} &\textbf{0.9330}&	0.8869&	0.8333	&0.5176\\
         \cline{3-5}
         && \multicolumn{3}{c|}{RaDE} &0.9256&	0.8806	&0.8475&	0.7243\\
         \cline{3-5}
           && \multirow{4}{*}{\rotatebox[origin=c]{90}{\textbf{GRaCE}}}& Rec. Density & Reciprocal &0.9258&	0.8874	&0.8447&	0.7433\\
           &&& Rec. Density & JacMax &0.9234&	0.8831	&0.8429	&0.7020\\
           &&& Acc. JacMax & Reciprocal &0.9258&	\textbf{0.8889}	&\textbf{0.8482}	&\textbf{0.7533}\\
          &&& Acc. JacMax & JacMax & 0.9211	&0.8816	&0.8421	&0.6877\\
          \cline{2-9}
          &\multirow{5}{*}{\rotatebox[origin=c]{45}{\textbf{DistilRoBERTa}}}& \multicolumn{3}{c|}{Original Features} &0.9258&	0.8840	&0.8344	&0.5396\\
         \cline{3-5}
         & &\multicolumn{3}{c|}{RaDE} &\textbf{0.9272}&	\textbf{0.8866}&\textbf{0.8476}	&0.6784\\
         \cline{3-5}
           && \multirow{4}{*}{\rotatebox[origin=c]{90}{\textbf{GRaCE}}}& Rec. Density & Reciprocal & 0.9225	&0.8812&	0.8445	&\textbf{0.7135}\\
           && &Rec. Density & JacMax & 0.9243&	0.8779	&0.8374	&0.6862\\
           && &Acc. JacMax & Reciprocal &0.9270&	0.8862	&0.8469	&0.6984\\
          && &Acc. JacMax & JacMax &0.9256	&0.8828	&0.8424	&0.6552\\
          \cline{2-9}
         &\multirow{6}{*}{\rotatebox[origin=c]{45}{\textbf{MiniLM}}}& \multicolumn{3}{c|}{Original Features} &0.9222&	0.8758&	0.8209	&0.5236\\
         \cline{3-5}
         && \multicolumn{3}{c|}{RaDE} &\textbf{0.9254}&	0.8815	&0.8409&	0.6428\\
         \cline{3-5}
           && \multirow{4}{*}{\rotatebox[origin=c]{90}{\textbf{GRaCE}}}& Rec. Density & Reciprocal &0.9220&	0.8828	&0.8408&	\textbf{0.7103}\\
           &&& Rec. Density & JacMax &0.9236&	0.8806	&0.8365	&0.6570\\
           &&& Acc. JacMax & Reciprocal &0.9238&	\textbf{0.8855}&\textbf{0.8426}&0.7091\\
          &&& Acc. JacMax & JacMax & 0.9229	&0.8810	&0.8398	&0.6342\\
          \hline
          \multirow{18}{*}{\rotatebox[origin=c]{45}{\textbf{WOS-5736}}} &\multirow{5}{*}{\rotatebox[origin=c]{45}{\textbf{mContriever}}}& \multicolumn{3}{c|}{Original Features} &0.8499&	0.7595&	0.6439	&0.2753\\
         \cline{3-5}
         && \multicolumn{3}{c|}{RaDE} &\textbf{0.8524}&	\textbf{0.7697}	&\textbf{0.6900}&	\textbf{0.4239}\\
         \cline{3-5}
           && \multirow{4}{*}{\rotatebox[origin=c]{90}{\textbf{GRaCE}}}& Rec. Density & Reciprocal &0.8399&	0.7492	&0.6610&	0.4154\\
           &&& Rec. Density & JacMax &0.8433&	0.7579	&0.6719	&0.4060\\
           && &Acc. JacMax & Reciprocal &0.8366&	0.7444	&0.6576	&0.4127\\
          & &&Acc. JacMax & JacMax & 0.8441	&0.7567	&0.6685	&0.3957\\
          \cline{2-9}
          &\multirow{6}{*}{\rotatebox[origin=c]{45}{\textbf{DistilRoBERTa}}}& \multicolumn{3}{c|}{Original Features} &\textbf{0.8479}&	0.7522	&0.6504&0.3481\\
         \cline{3-5}
         & &\multicolumn{3}{c|}{RaDE} &0.8449&	\textbf{0.7531}&\textbf{0.6665}	&0.3745\\
         \cline{3-5}
           && \multirow{4}{*}{\rotatebox[origin=c]{90}{\textbf{GRaCE}}}& Rec. Density & Reciprocal & 0.8301&0.7354&	0.6431	&0.3868\\
           & &&Rec. Density & JacMax & 0.8353&	0.7415	&0.6542	&0.3693\\
           && &Acc. JacMax & Reciprocal &0.8332&	0.7341	&0.6466	&\textbf{0.3931}\\
          &&& Acc. JacMax & JacMax &0.8350	&0.7424	&0.654	&0.3628\\
          \cline{2-9}
         &\multirow{5}{*}{\rotatebox[origin=c]{45}{\textbf{MiniLM}}}& \multicolumn{3}{c|}{Original Features} &\textbf{0.8489}&	\textbf{0.7555}&	0.6491	&0.3469\\
         \cline{3-5}
         && \multicolumn{3}{c|}{RaDE} &0.8406&	0.7531	&\textbf{0.6678}&	0.3903\\
         \cline{3-5}
           && \multirow{4}{*}{\rotatebox[origin=c]{90}{\textbf{GRaCE}}}& Rec. Density & Reciprocal &0.8304&	0.7349	&0.6461&	0.3946\\
           &&& Rec. Density & JacMax &0.8364&	0.7482	&0.6591	&0.3820\\
           & &&Acc. JacMax & Reciprocal &0.8319&	0.7406&0.6570&\textbf{0.4052}\\
          && &Acc. JacMax & JacMax & 0.8357	&0.7474	&0.6582	&0.3769\\
          \hline
    \end{tabular}
    }
    \label{tab:retrieval}
\end{table}

The retrieval results presented in Table \ref{tab:retrieval} further validate the superiority of GRaCE over the baseline methods, including RaDE and Original Features, while also demonstrating their competitiveness. In general, we observe that interpretable embedding methods, RaDE and GRaCE, tend to perform better at increasing retrieval depths. However, Original Features still exhibit strong performance across several datasets and descriptors, particularly at lower precision levels, often achieving the best results in P@2 and P@4.  A particularly notable trend is the consistent improvement in MAP values when using GRaCE. In fact, GRaCE achieved the highest MAP scores in 9 out of 12 cases, highlighting its robustness and ability to generalize across different retrieval tasks. This improvement is especially evident in the BBCNews dataset, where the MAP score jumps from around 50\% to over 70\% across all descriptors, demonstrating the effectiveness of GRaCE in enhancing retrieval performance at deeper ranking levels.

\subsection{Classification Tasks}
\label{subsec:classification-tasks}

\begin{table}[t!]
    \centering
    \caption{Evaluation of the Accuracy in classification methods.}
    \vspace{-2mm}
    \resizebox{.50\textwidth}{!}{
    \begin{tabular}{c|c|c|c|c|cccc}
    \hline
         \textbf{Dataset} & \textbf{Descriptor} & \multicolumn{2}{c|}{\textbf{QPP}} & \textbf{Correlation} & \textbf{KNN} & \textbf{NET} & \textbf{SGC} & \textbf{APPNP}\\
         \hline
          \multirow{18}{*}{\rotatebox[origin=c]{45}{\textbf{Flowers}}} & \multirow{6}{*}{\rotatebox[origin=c]{45}{\textbf{ViT-B16}}}& \multicolumn{3}{c|}{Original Features} &0.9543& 0.9282        & 0.9279        & 0.8985\\
         \cline{3-5}
         && \multicolumn{3}{c|}{RaDE} &\textbf{0.9781} & 0.9662        & \textbf{0.9692}        & \textbf{0.9710}\\
         \cline{3-5}
           && \multirow{4}{*}{\rotatebox[origin=c]{90}{\textbf{GRaCE}}}& Rec. Density & Reciprocal &0.9760 & \textbf{0.9738}        & 0.9547        & 0.9698\\
           &&& Rec. Density & JacMax &  0.9679& 0.9668        & 0.9597        & 0.9696\\
           &&& Acc. JacMax & Reciprocal &  0.9766& 0.9737        & 0.9443        & 0.9681\\
          &&& Acc. JacMax & JacMax & 0.9692& 0.9665        & 0.9597        & 0.9696\\
          \cline{2-9}
            &       \multirow{6}{*}{\rotatebox[origin=c]{45}{\textbf{Swin-Tf}}}& \multicolumn{3}{c|}{Original Features} & 0.9506& 0.9717        & 0.9701        & 0.9746\\
         \cline{3-5}
         & &\multicolumn{3}{c|}{RaDE} & 0.9889& \textbf{0.9947}        & \textbf{0.9951}        & \textbf{0.9953}\\
         \cline{3-5}
           && \multirow{4}{*}{\rotatebox[origin=c]{90}{\textbf{GRaCE}}}& Rec. Density & Reciprocal &0.9785& 0.9925        & 0.9774        & 0.9714\\
           &&& Rec. Density & JacMax &  \textbf{0.9906}& 0.9926        & 0.9680         & 0.9931\\
           &&& Acc. JacMax & Reciprocal &  0.9777& 0.9933        & 0.9783        & 0.9755\\
          & &&Acc. JacMax & JacMax & 0.9901& 0.9922        & 0.9667        & 0.9937\\
          \cline{2-9}
             &      \multirow{6}{*}{\rotatebox[origin=c]{45}{\textbf{DINOv2-B14}}}& \multicolumn{3}{c|}{Original Features} & 0.9941 & 0.9981        & 0.9977        & 0.9982\\
         \cline{3-5}
         & &\multicolumn{3}{c|}{RaDE} & 0.9965& 0.9976        & \textbf{0.9981}        & 0.9974\\
         \cline{3-5}
           && \multirow{4}{*}{\rotatebox[origin=c]{90}{\textbf{GRaCE}}}& Rec. Density & Reciprocal &0.9963 & \textbf{0.9983}        & 0.9886        & \textbf{0.9987}\\
           &&& Rec. Density & JacMax &  0.9959 & 0.9952        & 0.9945        & 0.9944\\
           &&& Acc. JacMax & Reciprocal &  0.9963 & 0.9977        & 0.9898        & 0.9956\\
          &&& Acc. JacMax & JacMax & \textbf{0.9977}& 0.9979        & 0.9953        & 0.9975 \\
          \hline
          \multirow{18}{*}{\rotatebox[origin=c]{45}{\textbf{Corel5k}}} & \multirow{6}{*}{\rotatebox[origin=c]{45}{\textbf{ViT-B16}}}& \multicolumn{3}{c|}{Original Features} &0.8756& \textbf{0.9242}        & \textbf{0.9336}        & 0.8679\\
         \cline{3-5}
         && \multicolumn{3}{c|}{RaDE} &\textbf{0.9211} & 0.9172        & 0.9234        & \textbf{0.9281}\\
         \cline{3-5}
           && \multirow{4}{*}{\rotatebox[origin=c]{90}{\textbf{GRaCE}}}& Rec. Density & Reciprocal &0.8955  & 0.8794        & 0.6761        & 0.7485\\
           &&& Rec. Density & JacMax &  0.8987 & 0.9006        & 0.7006        & 0.8679\\
           &&& Acc. JacMax & Reciprocal &  0.8953& 0.879         & 0.6778        & 0.7466\\
          & &&Acc. JacMax & JacMax & 0.9004 & 0.9061        & 0.710         & 0.8788\\
          \cline{2-9}
             &      \multirow{6}{*}{\rotatebox[origin=c]{45}{\textbf{Swin-Tf}}}& \multicolumn{3}{c|}{Original Features} & 0.8743 & \textbf{0.9571}        & \textbf{0.9575}        & \textbf{0.9631}\\
         \cline{3-5}
         & &\multicolumn{3}{c|}{RaDE} & \textbf{0.9407 }& 0.9424        & 0.9539        & 0.9499\\
         \cline{3-5}
           && \multirow{4}{*}{\rotatebox[origin=c]{90}{\textbf{GRaCE}}}& Rec. Density & Reciprocal &0.9023& 0.9136        & 0.7179        & 0.7983\\
           &&& Rec. Density & JacMax &  0.9094 & 0.9348        & 0.7222        & 0.9212\\
           &&& Acc. JacMax & Reciprocal &  0.9016 & 0.9101        & 0.7187        & 0.8118\\
          && &Acc. JacMax & JacMax & 0.9212 & 0.9421        & 0.7335        & 0.9329\\
          \cline{2-9}
             &      \multirow{6}{*}{\rotatebox[origin=c]{45}{\textbf{DINOv2-B14}}}& \multicolumn{3}{c|}{Original Features} & 0.9121& 0.9322        & 0.9329        & \textbf{0.9468}\\
         \cline{3-5}
         & &\multicolumn{3}{c|}{RaDE} & 0.9256 & 0.9316        & \textbf{0.9420}         & 0.9455\\
         \cline{3-5}
           && \multirow{4}{*}{\rotatebox[origin=c]{90}{\textbf{GRaCE}}}& Rec. Density & Reciprocal &0.9221 & 0.9249        & 0.7424        & 0.8305\\
           &&& Rec. Density & JacMax &  0.9286 & 0.9376        & 0.7807        & 0.9226\\
           &&& Acc. JacMax & Reciprocal &  \textbf{0.9313}& 0.9228        & 0.7324        & 0.8339\\
          & &&Acc. JacMax & JacMax & 0.9286 & \textbf{0.9431}        & 0.8007        & 0.9264\\
          \hline
          \multirow{18}{*}{\rotatebox[origin=c]{45}{\textbf{BBCNews}}} & \multirow{6}{*}{\rotatebox[origin=c]{45}{\textbf{mContriever}}}& \multicolumn{3}{c|}{Original Features} &0.8651 & \textbf{0.8218}        & \textbf{0.8325}        & \textbf{0.8271}\\
         \cline{3-5}
         & &\multicolumn{3}{c|}{RaDE} &\textbf{0.8812 } & 0.8176        & 0.8194        & 0.8207\\
         \cline{3-5}
           && \multirow{4}{*}{\rotatebox[origin=c]{90}{\textbf{GRaCE}}}& Rec. Density & Reciprocal &0.8755  & 0.8157        & 0.8062        & 0.8148\\
           &&& Rec. Density & JacMax &  0.8732 & 0.8167        & 0.8145        & 0.8153\\
           &&& Acc. JacMax & Reciprocal & 0.8796 & 0.8184        & 0.8209        & 0.8200\\
          & &&Acc. JacMax & JacMax & 0.8753 & 0.8185        & 0.8144        & 0.8157\\
          \cline{2-9}
             &      \multirow{6}{*}{\rotatebox[origin=c]{45}{\textbf{DistilRoBERTa}}}& \multicolumn{3}{c|}{Original Features} & 0.8695 	& 0.8202        & 0.8128        & 0.8161\\
         \cline{3-5}
         & &\multicolumn{3}{c|}{RaDE} & \textbf{0.8831}& 0.8171        & \textbf{0.8194}        & \textbf{0.8194}\\
         \cline{3-5}
           && \multirow{4}{*}{\rotatebox[origin=c]{90}{\textbf{GRaCE}}}& Rec. Density & Reciprocal &0.8781 & 0.8169        & 0.8075        & 0.8150\\
           &&& Rec. Density & JacMax &  0.8775& 0.8140         & 0.8107        & 0.8138 \\
           &&& Acc. JacMax & Reciprocal &  0.8818 & 0.8206        & 0.7924        & 0.8154\\
          && &Acc. JacMax & JacMax & 0.8816& \textbf{0.8214}        & \textbf{0.8194}        & 0.8191\\
          \cline{2-9}
             &      \multirow{6}{*}{\rotatebox[origin=c]{45}{\textbf{MiniLM}}}& \multicolumn{3}{c|}{Original Features} & 0.8679 & 0.8172        & 0.8104        & 0.8175\\
         \cline{3-5}
         & &\multicolumn{3}{c|}{RaDE} & \textbf{0.8786}& 0.8204        & \textbf{0.8237}        & \textbf{0.8188}\\
         \cline{3-5}
           && \multirow{4}{*}{\rotatebox[origin=c]{90}{\textbf{GRaCE}}}& Rec. Density & Reciprocal &0.8772& 0.8224        & 0.8029        & 0.8159\\
           &&& Rec. Density & JacMax &  0.8762& 0.8167        & 0.8140         & 0.8146\\
           &&& Acc. JacMax & Reciprocal &  \textbf{0.8786}& \textbf{0.8245}        & 0.7975        & 0.8162\\
          & &&Acc. JacMax & JacMax & 0.8784& 0.8174        & 0.8134        & 0.8141\\
          \hline
          \multirow{18}{*}{\rotatebox[origin=c]{45}{\textbf{WOS-5736}}} & \multirow{6}{*}{\rotatebox[origin=c]{45}{\textbf{mContriever}}}& \multicolumn{3}{c|}{Original Features} &0.6692& 0.7048         & 0.7013         & 0.6883\\
         \cline{3-5}
         & &\multicolumn{3}{c|}{RaDE} &\textbf{0.7156} & \textbf{0.7127}         & \textbf{0.7184}         & \textbf{0.7177}\\
         \cline{3-5}
           && \multirow{4}{*}{\rotatebox[origin=c]{90}{\textbf{GRaCE}}}& Rec. Density & Reciprocal &0.6912 & 0.6690          & 0.4460          & 0.5828\\
           &&& Rec. Density & JacMax &  0.7044 & 0.7081         & 0.6426         & 0.6914\\
           &&& Acc. JacMax & Reciprocal &  0.6893 & 0.6612         & 0.4363         & 0.5947\\
          & &&Acc. JacMax & JacMax & 0.7040 & 0.7122         & 0.6281         & 0.6930 \\
          \cline{2-9}
             &      \multirow{6}{*}{\rotatebox[origin=c]{45}{\textbf{DistilRoBERTa}}}& \multicolumn{3}{c|}{Original Features} & 0.6732& \textbf{0.6981}         & 0.6530          & 0.6367\\
         \cline{3-5}
         & &\multicolumn{3}{c|}{RaDE} & \textbf{0.6870}& 0.6782         & \textbf{0.6918}         & \textbf{0.6954}\\
         \cline{3-5}
           && \multirow{4}{*}{\rotatebox[origin=c]{90}{\textbf{GRaCE}}}& Rec. Density & Reciprocal &0.6599 & 0.6558         & 0.4734         & 0.6044\\
           &&& Rec. Density & JacMax &  0.6769 & 0.6845         & 0.6195         & 0.6714 \\
           &&& Acc. JacMax & Reciprocal &   0.6654 & 0.6577         & 0.4374         & 0.6065\\
          & &&Acc. JacMax & JacMax & 0.6796 & 0.6835         & 0.6114         & 0.6764\\
          \cline{2-9}
             &      \multirow{6}{*}{\rotatebox[origin=c]{45}{\textbf{MiniLM}}}& \multicolumn{3}{c|}{Original Features} & 0.6751 & \textbf{0.7028}        & 0.6619         & 0.6362\\
         \cline{3-5}
         & &\multicolumn{3}{c|}{RaDE} & \textbf{0.6919}& 0.6763         & \textbf{0.6934}         & \textbf{0.6978}\\
         \cline{3-5}
           && \multirow{4}{*}{\rotatebox[origin=c]{90}{\textbf{GRaCE}}}& Rec. Density & Reciprocal &0.6662& 0.6657         & 0.4917         & 0.6128\\
           &&& Rec. Density & JacMax &  0.6827 & 0.6867         & 0.6257         & 0.6733 \\
           &&& Acc. JacMax & Reciprocal &  0.6789 & 0.6766         & 0.4646         & 0.6142\\
          && &Acc. JacMax & JacMax & 0.6833 & 0.6862         & 0.6229         & 0.6758\\
          \hline
    \end{tabular}
    }
    \label{tab:classification}
\end{table}

The classification results are presented in Table \ref{tab:classification}. Across the analyzed datasets and feature descriptors, graph embedding methods struggle to consistently outperform the results obtained using Original Features. Even in cases where RaDE and GRaCE achieve the best performance, the improvements are generally marginal, particularly when examining the results of GCN models. However, the KNN classifier demonstrates significant benefits from the embedding methods, consistently outperforming the Original Features across all cases. Notable improvements include the Flowers dataset with the Swin-Tf descriptor, where RaDE with the Reciprocal Density and JaccardMax combination enhances Accuracy by 4\% compared to Original Features. Similarly, in the Corel5k dataset using the Swin-Tf descriptor and RaDE embeddings, classification Accuracy improves by nearly 7\%, underscoring the effectiveness of these embedding methods in specific scenarios. These results reinforce the advantages of interpretable embedding methods in enhancing classification performance across diverse datasets and feature extraction techniques.

\subsection{Clustering Tasks}
\label{subsec:clustering-tasks}

\begin{table}[t!]
    \centering
        \caption{Evaluation of NMI in clustering algorithms.}
\vspace{-2mm}
    \resizebox{.50\textwidth}{!}{
    \begin{tabular}{c|c|c|c|c|cccc}
    \hline
         \textbf{Dataset} & \textbf{Descriptor} & \multicolumn{2}{c|}{\textbf{QPP}} & \textbf{Correlation} & \textbf{HDBSCAN} & \textbf{AP} & \textbf{KMeans} & \textbf{AHC}\\
          \hline
         \multirow{18}{*}{\rotatebox[origin=c]{45}{\textbf{Flowers}}} & \multirow{6}{*}{\rotatebox[origin=c]{45}{\textbf{ViT-B16}}}& \multicolumn{3}{c|}{Original Features} & 0.9082 & 0.8260&0.9302&0.9160\\
         \cline{3-5}
         && \multicolumn{3}{c|}{RaDE} &\textbf{0.9382}&\textbf{0.8640}&0.9610&\textbf{0.9644}\\
         \cline{3-5}
           && \multirow{4}{*}{\rotatebox[origin=c]{90}{\textbf{GRaCE}}}& Rec. Density & Reciprocal &0.8705&0.6760&0.9620&0.9604\\
           &&& Rec. Density & JacMax &0.8592&0.7963&0.9512&0.9479\\
           &&& Acc. JacMax & Reciprocal &0.8587&0.6283&\textbf{0.9622}&0.9606\\
          &&& Acc. JacMax & JacMax &0.8677&0.8124&0.9497&0.9470\\
          \cline{2-9}
          &\multirow{6}{*}{\rotatebox[origin=c]{45}{\textbf{Swin-Tf}}}& \multicolumn{3}{c|}{Original Features} & 0.9354 & \textbf{0.8420}&0.9387&0.9895\\
         \cline{3-5}
         & &\multicolumn{3}{c|}{RaDE} &\textbf{0.9853}&0.7490&\textbf{0.9872}&\textbf{0.9910}\\
         \cline{3-5}
           && \multirow{4}{*}{\rotatebox[origin=c]{90}{\textbf{GRaCE}}}& Rec. Density & Reciprocal &0.9199&0.6292&0.9820&\textbf{0.9910}\\
           && &Rec. Density & JacMax &0.9561&0.6585&0.9863&0.9837\\
           && &Acc. JacMax & Reciprocal &0.9294&0.6148&0.9816&\textbf{0.9910}\\
          & &&Acc. JacMax & JacMax &0.9431&0.6524&0.9863&0.9831\\
          \cline{2-9}
          &\multirow{6}{*}{\rotatebox[origin=c]{45}{\textbf{DINOv2-B14}}}& \multicolumn{3}{c|}{Original Features} & 0.9630 & \textbf{0.8985}&0.9813&0.9986\\
         \cline{3-5}
         & &\multicolumn{3}{c|}{RaDE} &\textbf{0.9796}&0.8826&0.9951&0.9951\\
         \cline{3-5}
           && \multirow{4}{*}{\rotatebox[origin=c]{90}{\textbf{GRaCE}}}& Rec. Density & Reciprocal &0.9718&0.6330&0.9953&0.9944\\
           & &&Rec. Density & JacMax &0.9370&0.6449&0.9951&0.9931\\
           &&& Acc. JacMax & Reciprocal &0.9503&0.6683&0.9945&0.9944\\
          & &&Acc. JacMax & JacMax &0.9573&0.6870&\textbf{0.9967}&\textbf{1}\\
           \hline
         \multirow{18}{*}{\rotatebox[origin=c]{45}{\textbf{Corel5k}}} & \multirow{6}{*}{\rotatebox[origin=c]{45}{\textbf{VIT B16}}}& \multicolumn{3}{c|}{Original Features} & 0.7340 & 0.8143&0.8938&\textbf{0.9255}\\
         \cline{3-5}
         && \multicolumn{3}{c|}{RaDE} &\textbf{0.8613}&\textbf{0.8590}&\textbf{0.9124}&0.9088\\
         \cline{3-5}
           && \multirow{4}{*}{\rotatebox[origin=c]{90}{\textbf{GRaCE}}}& Rec. Density & Reciprocal &0.8196&0.7896&0.9048&0.9064\\
           &&& Rec. Density & JacMax &0.8161&0.7847&0.8989&0.8959\\
           &&& Acc. JacMax & Reciprocal &0.8143&0.7965&0.8943&0.8959\\
          &&& Acc. JacMax & JacMax &0.7776&0.8322&0.8922&0.8922\\
          \cline{2-9}
          &\multirow{6}{*}{\rotatebox[origin=c]{45}{\textbf{Swin Tf}}}& \multicolumn{3}{c|}{Original Features} & 0.7148 & 0.8169&\textbf{0.9359}&\textbf{0.9712}\\
         \cline{3-5}
         & &\multicolumn{3}{c|}{RaDE} &\textbf{0.8876}&\textbf{0.8504}&0.9353&0.9425\\
         \cline{3-5}
           && \multirow{4}{*}{\rotatebox[origin=c]{90}{\textbf{GRaCE}}}& Rec. Density & Reciprocal &0.8567&0.8021&0.9108&0.9288\\
           & &&Rec. Density & JacMax &0.8510&0.7915&0.9039&0.9226\\
           & &&Acc. JacMax & Reciprocal &0.8488&0.7866&0.9132&0.9293\\
          && &Acc. JacMax & JacMax &0.8399&0.7908&0.9025&0.9254\\
          \cline{2-9}
          &\multirow{6}{*}{\rotatebox[origin=c]{45}{\textbf{DINOv2 B14}}}& \multicolumn{3}{c|}{Original Features} & 0.7912 & 0.8021&0.9243&0.9362\\
         \cline{3-5}
         & &\multicolumn{3}{c|}{RaDE} &0.8396&0.8376&0.9237&0.9340\\
         \cline{3-5}
           && \multirow{4}{*}{\rotatebox[origin=c]{90}{\textbf{GRaCE}}}& Rec. Density & Reciprocal &\textbf{0.8594}&0.8288&\textbf{0.9353}&0.9323\\
           &&& Rec. Density & JacMax &0.7989&\textbf{0.8492}&0.9229&0.9282\\
           &&& Acc. JacMax & Reciprocal &0.8301&0.8119&0.9339&\textbf{0.9395}\\
          &&& Acc. JacMax & JacMax &0.7871&0.8249&0.9218&0.9334\\
          \hline
          \multirow{18}{*}{\rotatebox[origin=c]{45}{\textbf{BBCNews}}} & \multirow{6}{*}{\rotatebox[origin=c]{45}{\textbf{mContriever}}}& \multicolumn{3}{c|}{Original Features} & 0.2740 & 0.3834&0.5999&\textbf{0.7387}\\
         \cline{3-5}
         & &\multicolumn{3}{c|}{RaDE} &0.4994&0.4818&\textbf{0.7153}&0.7309\\
         \cline{3-5}
           && \multirow{4}{*}{\rotatebox[origin=c]{90}{\textbf{GRaCE}}}& Rec. Density & Reciprocal &\textbf{0.5526}&\textbf{0.5293}&0.6920&0.6551\\
           & &&Rec. Density & JacMax &0.4683&0.4691&0.6727&0.6105\\
           && &Acc. JacMax & Reciprocal &0.5455&0.4434&0.6166&0.7045\\
          & &&Acc. JacMax & JacMax &0.4686&0.4662&0.5854&0.6338\\
          \cline{2-9}
         &\multirow{6}{*}{\rotatebox[origin=c]{45}{\textbf{DistilRoBERTa}}}& \multicolumn{3}{c|}{Original Features} & 0.0521 & 0.3961&0.5992&\textbf{0.7111}\\
         \cline{3-5}
         && \multicolumn{3}{c|}{RaDE} &0.4747&0.4720&0.6358&0.5654\\
         \cline{3-5}
           && \multirow{4}{*}{\rotatebox[origin=c]{90}{\textbf{GRaCE}}}& Rec. Density & Reciprocal &0.4939&0.4478&\textbf{0.6424}&0.6294\\
           &&& Rec. Density & JacMax &\textbf{0.5299}&0.4741&0.5932&0.5265\\
           &&& Acc. JacMax & Reciprocal &0.4588&\textbf{0.4842}&0.5206&0.5178\\
          &&& Acc. JacMax & JacMax &0.4456&0.4543&0.5272&0.4886\\
          \cline{2-9}
          &\multirow{6}{*}{\rotatebox[origin=c]{45}{\textbf{MiniLM}}}& \multicolumn{3}{c|}{Original Features} & 0.2752 & 0.3814&0.5794&\textbf{0.6785}\\
         \cline{3-5}
         & &\multicolumn{3}{c|}{RaDE} &0.5010&0.4658&0.6188&0.6576\\
         \cline{3-5}
           && \multirow{4}{*}{\rotatebox[origin=c]{90}{\textbf{GRaCE}}}& Rec. Density & Reciprocal &\textbf{0.5744}&0.4993&\textbf{0.7159}&0.6139\\
           && &Rec. Density & JacMax &0.3285&0.4650&0.5786&0.6417\\
           && &Acc. JacMax & Reciprocal &0.5880&\textbf{0.5068}&0.5999&0.6339\\
          & &&Acc. JacMax & JacMax &0.3161&0.4659&0.5177&0.6068\\
          \hline
          \multirow{18}{*}{\rotatebox[origin=c]{45}{\textbf{WOS-5736}}} & \multirow{6}{*}{\rotatebox[origin=c]{45}{\textbf{mContriever}}}& \multicolumn{3}{c|}{Original Features} & 0.1140 & 0.3974&0.5411&0.4976\\
         \cline{3-5}
         && \multicolumn{3}{c|}{RaDE} &0.4077&0.4638&\textbf{0.5424}&0.5317\\
         \cline{3-5}
           && \multirow{4}{*}{\rotatebox[origin=c]{90}{\textbf{GRaCE}}}& Rec. Density & Reciprocal &0.4358&\textbf{0.4753}&0.5330&\textbf{0.5436}\\
           &&& Rec. Density & JacMax &0.0167&0.4619&0.5392&0.5177\\
           &&& Acc. JacMax & Reciprocal &\textbf{0.4376}&0.4642&0.5400&0.5226\\
          &&& Acc. JacMax & JacMax &0.0086&0.4550&0.5353&0.5385\\
          \cline{2-9}
         &\multirow{6}{*}{\rotatebox[origin=c]{45}{\textbf{DistilRoBERTa}}}& \multicolumn{3}{c|}{Original Features} &0.1547&0.4066&\textbf{0.5301}&0.4939\\
         \cline{3-5}
         && \multicolumn{3}{c|}{RaDE} &0.4005&0.4468&0.5264&0.5185\\
         \cline{3-5}
           && \multirow{4}{*}{\rotatebox[origin=c]{90}{\textbf{GRaCE}}}& Rec. Density & Reciprocal &\textbf{0.4031}&0.4453&0.5211&0.5230\\
           &&& Rec. Density & JacMax &0.3729&0.4438&0.4960&\textbf{0.5274}\\
           & &&Acc. JacMax & Reciprocal &0.4002&\textbf{0.4632}&0.5194&0.5201\\
          & &&Acc. JacMax & JacMax &0.0077&0.4421&0.5040&0.4967\\
          \cline{2-9}
          &\multirow{6}{*}{\rotatebox[origin=c]{45}{\textbf{MiniLM}}}& \multicolumn{3}{c|}{Original Features} & 0.0453 & 0.4066&0.5280&0.5076\\
         \cline{3-5}
         & &\multicolumn{3}{c|}{RaDE} &0.3503&0.4536&0.5213&0.5145\\
         \cline{3-5}
           && \multirow{4}{*}{\rotatebox[origin=c]{90}{\textbf{GRaCE}}}& Rec. Density & Reciprocal &0.3907&\textbf{0.4648}&0.5160&\textbf{0.5275}\\
           &&& Rec. Density & JacMax &0.0086&0.4516&0.5071&0.5196\\
           &&& Acc. JacMax & Reciprocal &\textbf{0.3958}&0.4608&0.5300&0.5072\\
          & &&Acc. JacMax & JacMax &0.0337&0.4499&\textbf{0.5332}&0.4986\\
          \hline
    \end{tabular}
    }
    \label{tab:clustering}
\end{table}

Table \ref{tab:clustering} presents the NMI scores for the four clustering algorithms evaluated across all datasets. As anticipated, graph embedding techniques generally improved the results compared to the Original Features. The most notable improvements were observed with HDBSCAN, where interpretable embeddings outperformed the Original Features in all cases. A striking example is the BBCNews dataset with DistilRoBERTa features, where the Original Features achieved an NMI score of 0.0521, while GRaCE significantly improved the result to 0.5299. This trend was consistent across all datasets for HDBSCAN. Conversely, the Agglomerative Hierarchical Clustering algorithm was the least affected by graph embeddings; however, GRaCE still managed to achieve a perfect NMI score of 1 on the Flowers dataset with DINOv2-B14 features. Affinity Propagation and KMeans also benefited from graph embeddings, particularly in textual datasets, while also showing improvements in image collections. Overall, clustering algorithms exhibited strong performance with graph embedding techniques, with RaDE excelling in image datasets and GRaCE standing out in textual data.

\section{Conclusion}
\label{sec:conclusion}
This study evaluates the GRaCE framework across multiple machine learning tasks, including subset selection, retrieval, classification, and clustering, using four datasets and six state-of-the-art feature descriptors, considering Transformer models for both textual and image datasets. GRaCE consistently demonstrates strong performance, often outperforming Original Features and proving to be a viable alternative to RaDE.

For subset selection, our experiments revealed that RaDE is an effective approach, achieving strong results in Clustering Accuracy and high Coverage across most evaluated cases. However, GRaCE matches or surpasses RaDE in many instances, demonstrating its effectiveness for possible subset selection in other tasks.

In retrieval tasks, GRaCE maintains a strong performance similar to RaDE. As noted in RaDE’s original publication, graph-based embeddings enhance retrieval tasks by improving performance at deeper ranking levels. GRaCE exhibits this same behavior, making it particularly valuable for retrieval scenarios requiring comprehensive evaluations. However, a slight reduction in precision at shallower depths suggests trade-offs that warrant further investigation.

Classification results reveal the benefits and limitations of graph embedding methods across different datasets and feature descriptors. While Original Features remain competitive in certain cases, both RaDE and GRaCE show significant improvements, particularly for textual datasets. The advantages of graph embeddings are particularly pronounced with the KNN classifier, where both RaDE and GRaCE outperform Original Features, emphasizing the potential of structured embeddings in classification tasks.

Clustering results further validate the advantages of graph-based embeddings, as they significantly enhance NMI scores across various clustering algorithms and datasets. RaDE shows strong performance in image datasets, whereas GRaCE excels in textual data, highlighting their complementary strengths.

Overall, GRaCE proves versatile across tasks, highlighting graph embeddings’ ability to capture meaningful data structures in a fully unsupervised way. RaDE and GRaCE complement each other, with GRaCE emerging as a robust option for subset selection and textual and multimedia applications. 

Future work should evaluate GRaCE against state-of-the-art graph embedding approaches and investigate the scalability and complexity of the proposed approach.

\section*{Acknowledgment}

The authors acknowledge the financial support from the São Paulo Research Foundation - FAPESP (grants \#2024/04890-5 and \#2023/00694-4), the Brazilian National Council for Scientific and Technological Development - CNPq (grants \#313193/2023-1 and \#422667/2021-8), Petrobras (grant \#2023/00095-3), and the University of São Paulo (PRPI Ordinance No. 1032). The authors are also grateful to FAPESP and the University of Manchester for their support in the context of the SPRINT program.

\bibliographystyle{ieeetr} 
\bibliography{references}

\end{document}